\documentclass{article}
\usepackage{ijcai24}

\usepackage{times}
\usepackage{soul}
\usepackage{url}
\usepackage[hidelinks]{hyperref}
\usepackage[utf8]{inputenc}
\usepackage[small]{caption}
\usepackage{graphicx}
\usepackage{amsmath}
\usepackage{amsthm}
\usepackage{booktabs}
\usepackage{algorithm}
\usepackage{algorithmic}
\usepackage[switch]{lineno}

\usepackage{makecell}
\usepackage{booktabs}  
\usepackage{threeparttable}
\usepackage{multirow}
\usepackage{epstopdf}
\usepackage{algorithm}
\usepackage{algorithmic}
\usepackage{amsfonts} 
\usepackage{epstopdf}
\usepackage{booktabs}
\usepackage{epsfig} 
\usepackage{graphicx}
\usepackage{amsmath}
\usepackage{amssymb, amsthm}
\usepackage{booktabs}

\usepackage{colortbl}

\usepackage{times}
\usepackage{epsfig}
\usepackage{graphicx}
\usepackage{amsmath}
\usepackage{amssymb}

\usepackage{float}
\usepackage{booktabs}
\usepackage{multicol}
\usepackage{multirow}
\usepackage{amsmath}
\usepackage{bm}
\usepackage{algorithm}
\usepackage{algorithmic}
\usepackage{xcolor}

\newcommand{\uptri}{\begingroup\color{red}\blacktriangle\endgroup}
\newcommand{\downtri}{\begingroup\color{cyan}\blacktriangledown\endgroup}

\newtheorem{theorem}{Theorem}

\title{Visual Imitation Learning with Calibrated Contrastive Representation}

\author{
Yunke Wang$^1$
\and
Linwei Tao$^2$
\and
Bo Du$^1$
\and
Yutian Lin$^1$
\And
Chang Xu$^2$
\affiliations
$^1$School of Computer Science, Wuhan University, China\\
$^2$School of Computer Science, Faculty of Engineering, The University of Sydney, Australia\\
\emails
\{yunke.wang, dubo, yutian.lin\}@whu.edu.cn,
\{linwei.tao, c.xu\}@sydney.edu.au
}

\begin{document}

\maketitle

\begin{abstract}
Adversarial Imitation Learning (AIL) allows the agent to reproduce expert behavior with low-dimensional states and actions. However, challenges arise in handling visual states due to their less distinguishable representation compared to low-dimensional proprioceptive features. While existing methods resort to adopt complex network architectures or separate the process of learning representation and decision-making, they overlook valuable intra-agent information within demonstrations. To address this problem, this paper proposes a simple and effective solution by incorporating calibrated contrastive representative learning into visual AIL framework. Specifically, we present an image encoder in visual AIL, utilizing a combination of unsupervised and supervised contrastive learning to extract valuable features from visual states. Based on the fact that the improved agent often produces demonstrations of varying quality, we propose to calibrate the contrastive loss by treating each agent demonstrations as a mixed sample. The incorporation of contrastive learning can be jointly optimized with the AIL framework, without modifying the architecture or incurring significant computational costs. Experimental results on DMControl Suite demonstrate our proposed method is sample efficient and can outperform other compared methods from different aspects.
\end{abstract}

\section{Introduction}
Imitation Learning (IL)~\cite{hussein2017imitation,zheng2022imitation} is a practical approach for addressing sequential decision-making problems~\cite{silver2016mastering,van2016deep}, which seeks to acquire an optimal policy for an agent by emulating the behavior of an expert. One of the primary approaches utilized in IL is Behavioral Cloning (BC)~\cite{pomerleau1988alvinn}, in which the agent observes the action of the expert and learns a straight mapping from state to action via regression. However, this offline training manner may suffer from compounding errors~\cite{brantley2019disagreement,xu2020error,tu2022sample} when the agent executes the policy, leading it to drift to new and dangerous states. Instead, Adversarial Imitation Learning (AIL)~\cite{ho2016generative,fu2017learning,wang2021learning} encourages the agent to cover the distribution of the expert policy, which can result in a more accurate policy. 
Despite the significant success of AIL in benchmark tasks~\cite{todorov2012mujoco}, where the expert demonstration is often of low dimension and represents the proprioceptive features (\textit{e.g.}, positions, velocities, and accelerations), its performance will degrade a lot in more challenging tasks with high-dimensional visual demonstrations~\cite{jaderberg2019human,espeholt2018impala,tucker2018inverse} (\textit{e.g.}, first-person camera observations). 
\begin{figure}[!t]
    \centering
    \includegraphics[width=3in]{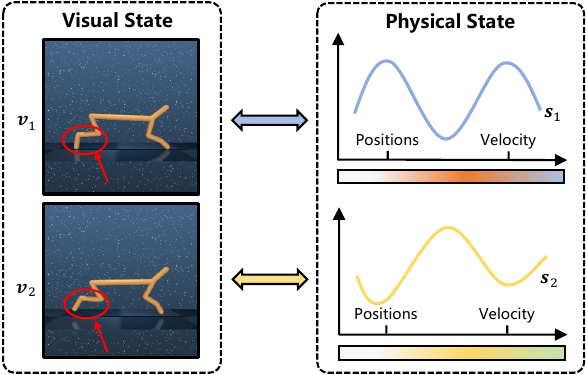}
    \caption{Two visual states $\textbf{v}_1, \textbf{v}_2$ and their corresponding physical states $\textbf{s}_1, \textbf{s}_2$ are shown in the figure. The physical state on the right column contains proprioceptive information (\textit{i.e.}, positions and velocities). Although a significant change in the physical state occurs, it may only result in slight changes to the visual state.}
    \label{fig:motivation}
\end{figure}

Recently, several studies have proposed solutions for visual imitation learning problems by enhancing the state representation~\cite{brown2019extrapolating,brown2020better,cai2021imitation,tucker2018inverse,yuan20183d,liu2023visual}. These approaches often require sophisticated training strategies that first learn a reward function or image encoder and prohibit end-to-end training~\cite{brown2019extrapolating,brown2020better}. Additionally, some of these methods are tailored to specific tasks, such as video games~\cite{cai2021imitation,tucker2018inverse,yuan20183d}. PatchAIL~\cite{liu2023visual}, on the other hand, adopts a different approach by evaluating the local expertise of various image patches rather than inducing a scalar reward from the entire image. By leveraging local information, PatchAIL can focus on minor motion changes at the patch level and obtain a more discriminative representation. PCIL~\cite{huang2023policy} adds constraints on the discriminator to push expert states together and pull away agent states to enhance its ability. However, it fails to leverage samples in the replay buffer and ignores the potential improvement of the agent during adversarial training.

Despite recent advances in visual-based adversarial imitation learning, the performance of such methods still lags behind that of proprioceptive-based adversarial imitation learning. The reason for this performance disparity can be attributed to the fact that proprioceptive features are highly distinguishable in low-dimensional feature space, as they contain explicit information about the agent. Conversely, certain features are not easily discernible in the image space. In Figure \ref{fig:motivation}, the significant change in the physical state only leads to minor changes in the visual state.
Hence, to enhance the performance of visual adversarial imitation learning, it is essential to develop a suitable representation that can accurately encode the visual states.
Contrastive representation learning is a technique that seeks to learn a representation space in which comparable instances are situated in close proximity while non-comparable instances are positioned far apart. This approach has demonstrated considerable potential in enhancing feature representations. Prior research in imitation learning has primarily focused on comparing expert and agent demonstrations, overlooking the intra-agent information present in the demonstrations. 

To address this limitation, we introduce our proposed method, named Contrastive Adversarial Imitation Learning (CAIL). Our approach aims to learn a feature representation that not only discriminates between expert and agent demonstrations, but also captures the similarities and differences among different agent demonstrations. Specifically, CAIL utilizes contrastive representation learning to encourage the representations of different agent demonstrations to be separated in the feature space while bringing similar demonstrations closer.
In practice, on top of the vanilla AIL, we add two contrastive losses on the image encoder. The first unsupervised contrastive loss makes the encoder fully exploit the larger amounts of agent demonstrations in the replay buffer. A supervised contrastive loss is then applied to distinguish agent and expert demonstrations while maintaining consistency with the unsupervised contrastive loss. 
Considering agent demonstrations improving with training, we regard the agent demonstration as a sample drawn from a mixture of agent and expert policy distribution and propose a calibrated supervised contrastive loss accordingly. Empirical results on DMControl Suite~\cite{tunyasuvunakool2020dm_control} show the contrastive learning combined AIL framework greatly outperforms compared methods.

\section{Related Work}
Reinforcement Learning (RL)~\cite{sutton2018reinforcement} aims to learn a policy for the agent by rewarding its action during its interaction with the environment. When the state is represented by raw pixels, it is essential to learn a good state representation through an image encoder to achieve effective results. In visual RL, it is common to share the latent representation between the policy and critic, with only the critic loss used for updating the encoder. Recent investigations~\cite{cobbe2019quantifying,lee2019network} into the effectiveness of data augmentation in pixel-based RL have shown that simple techniques, such as cutout and random convolutional, can enhance the generalization of agents across different visual RL benchmarks. CURL~\cite{laskin2020curl} leverages data augmentations to learn a contrastive representation in the RL setting, which enhances the data-efficiency of pixel-based RL. Similarly, some other methods~\cite{finn2015learning,yarats2021improving,yarats2021reinforcement,kostrikov2020image} consider representation learning as an auxiliary task and train a regularized autoencoder jointly with the RL agent. RAD~\cite{laskin2020reinforcement} utilizes data augmentation directly without introducing auxiliary loss, achieving good data-efficient results in pixel-based RL. 

Based on visual reinforcement learning, visual imitation learning enables the agent to learn a policy from visual demonstrations. This approach has become increasingly popular due to its compatibility with real-world tasks such as autonomous driving and robot learning~\cite{wen2021keyframe,ross2011reduction}.
Similar to visual reinforcement learning, a good image encoder is essential to achieve good results in visual imitation learning as it is difficult to learn effective representations in unstable adversarial training. To address this problem, T-REX~\cite{brown2019extrapolating} and D-REX~\cite{brown2020better} extrapolate a reward function from visual observations via ranked trajectories, then learn the agent policy using the learned reward function. 
The first work to successfully apply AIL to video game environments such as Atari games is \cite{tucker2018inverse}, which also conducts pre-training on the encoder first. Some recent works~\cite{rafailov2021visual,torabi2018behavioral,cohen2021imitation,liu2023visual} have directly modified plain adversarial imitation learning methods to make them compatible with visual inputs, either through network architecture or training strategies. 
However, there is still a clear gap between visual AIL and plain AIL.

\section{Preliminary}
\noindent\textbf{Markov Decision Process (MDP).}
MDP is popular to formulate reinforcement learning (RL) \cite{DBLP:books/wi/Puterman94} and imitation learning (IL) problems. An MDP normally consists six basic elements $\mathcal{M}=(\mathcal{S}, \mathcal{A}, \mathcal{P}, \mathcal{R}, \gamma, \mu_0)$, where $\mathcal{S}$ is a set of states, $\mathcal{A}$ is a set of actions, $\mathcal{P}:\mathcal{S}\times\mathcal{A}\times\mathcal{S}\rightarrow [0, 1]$ is the stochastic transition probability from current state $s$ to the next state $s'$, $\mathcal{R}:\mathcal{S}\times\mathcal{A}\rightarrow \mathbb{R}$ is the obtained reward of agent when taking action $a$ in a certain state $s$, $\gamma \in [0,1]$ is the discounted rate and $\mu_0$ denotes the initial state distribution. Given a trajectory $\tau  = \{(s_t, a_t)\}_{t=0}^{T}$, the return $R(\tau)$ is defined as the discounted sum of rewards obtained by the agent over this episode, $R(\tau)=\sum_{t=0}^{T} \gamma^k r(s_k, a_k)$ and $T$ is the number of steps to reach an absorbing state. The goal of RL is thus to learn a policy that can maximize the expected return over all episodes during the interaction. For any policy $\pi:\mathcal{S}\rightarrow\mathcal{A}$, there is an one-to-one correspondence between $\pi$ and its occupancy measure $\rho_\pi:\mathcal{S}\times\mathcal{A}\rightarrow[0, 1]$.

\noindent\textbf{Adversarial Imitation Learning (AIL).}
Generative Adversarial Imitation Learning (GAIL)~\cite{ho2016generative} is the most representative work, which directly applies the general GAN framework~\cite{goodfellow2014generative} into adversarial imitation learning. Given a set of expert demonstrations $(s,a)$ drawn from the expert policy $\pi_e$, GAIL aims to learn an agent policy $\pi_\theta$ by minimizing the Jensen-Shannon divergence between $\rho_{\pi_\theta}$ and $\rho_{\pi_e}$~\cite{ke2020imitation}. 
In the implementation, a discriminator $D_\phi$ is introduced to distinguish demonstrations from expert and agent policy, yet the agent policy tries its best to `fool' the discriminator:
\begin{align}
    \label{gail_d}
    \min_\theta\max_\phi \ & \mathbb{E}_{(s,a)\sim\rho_{\pi_e}} [\log D_\phi(s,a)] \\ \nonumber & + \mathbb{E}_{(s,a)\sim\rho_{\pi_\theta}} [\log(1-D_\phi(s,a))].
\end{align}
The agent is trained to minimize the outer objective function $\mathbb{E}_{(s,a)\sim\rho_{\pi_\theta}} [\log(1-D_\phi(s,a))]$, and therefore the output of $-\log(1-D(s,a))$ can be regarded as reward. Regular RL methods like TRPO \cite{schulman2015trust}, PPO~\cite{schulman2017proximal} and SAC \cite{haarnoja2018soft} can be thus used to update the agent policy $\pi_\theta$. 

Typically, in the visual imitation learning problem, visual demonstrations usually do not contain expert actions. We therefore replace the state-action pair with the visual state here and denote it as $\bm v$ to avoid confusion. With an image encoder $f: \bm v\rightarrow \mathbb R^{dim(r)}$ and an MLP projection head $h_d: \mathbb R^{dim(r)}\rightarrow \mathbb R^1$, where $dim(r)$ denotes the dimension of the extracted representation $r$ we can re-formulate the discriminator $D\rightarrow h_d\circ f$. The objective of training discriminator can be rewritten as,
\begin{align}
    \label{gail_df}
    \mathcal{L}(h_d,f)=-& \mathbb{E}_{\bm v\sim\rho_{\pi_e}} [\log h_d(f(\bm v))] \\ \nonumber &-\mathbb{E}_{\bm v\sim\rho_{\pi_\theta}} [\log(1-h_d(f(\bm v)))],
\end{align}
The agent policy $\pi_\theta$ can then be trained with reward $-\log(1-h_d(f(\bm v)))$ via reinforcement learning.

\begin{figure*}[!thp]
    \centering    
    \includegraphics[width=0.95\textwidth]{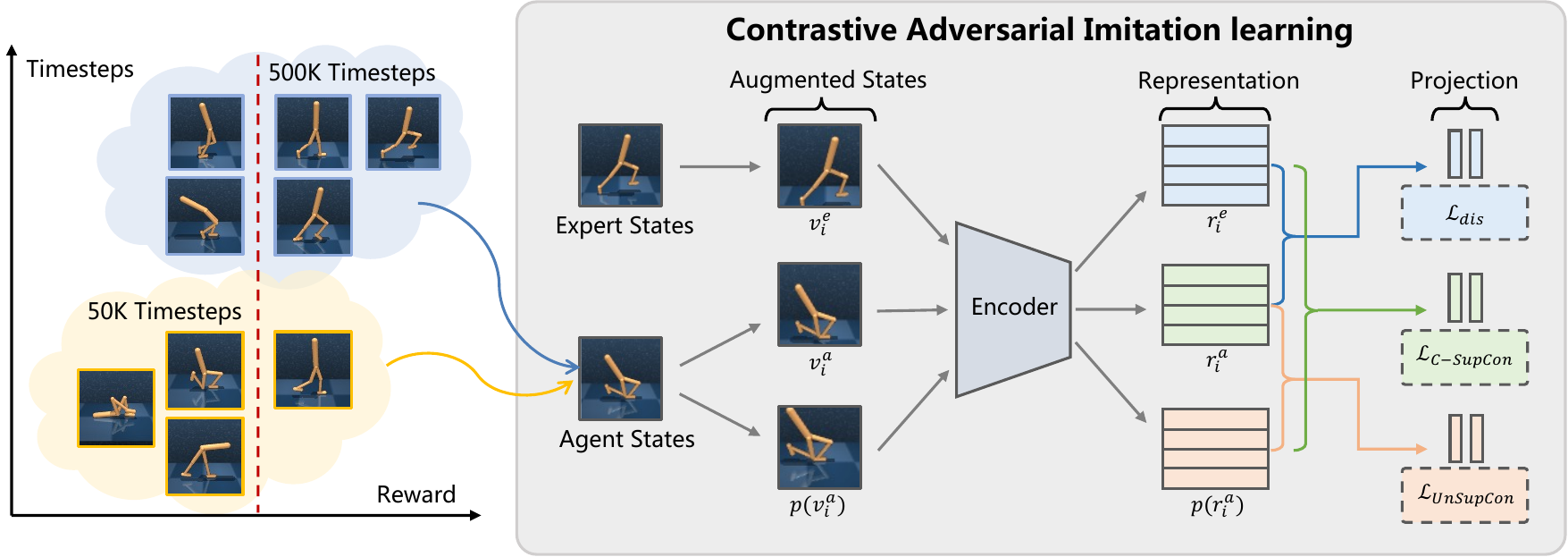}
    \caption{An overview of Contrastive Adversarial Imitation Learning. The encoder extracts the representation of the augmented expert state, and two augmented agent states. The training objective of the encoder consists of a discrimination loss, an unsupervised contrastive loss and a calibrated supervised contrastive loss.}
    \label{fig:CAIL}
\end{figure*}

\section{Methodology}
Applying GAIL (Eq. (\ref{gail_df})) directly to visual imitation learning may result in a decrease in performance, primarily due to the limited discriminative capacity of the encoder with respect to visually-similar yet semantically-distinct states. Thus, a more stringent criterion for the representation ability of the encoder is necessary. However, the instability associated with adversarial training and random sampling in RL poses obstacles to effectively learning a robust encoder. In recent times, contrastive learning has emerged as a potent methodology for extracting discriminative representations. The fundamental concept behind integrating contrastive learning into GAIL involves deriving an informative representation $\bm r$ of states $\bm v$ by training an encoder function $f$ that brings the representations $\bm r = f(\bm v)$ of semantically similar states (positives) closer to each other while moving those of dissimilar states (negatives) further apart in the embedding space. 

To this end, we propose a novel training scheme, named CAIL, that incorporates contrastive learning as a fundamental training objective. To begin with, we explore the adoption of an unsupervised contrastive loss function denoted as $\mathcal{L_{\text{UnSupCon}}}$, which serves to fully exploit the agent visual states. Additionally, we incorporate a supervised contrastive loss function denoted as $\mathcal{L_{\text{SupCon}}}$, to enhance the encoder's discriminative ability. Motivated by the observation that agent states may learn the expert characteristics to such a high degree that it is no longer appropriate to treat them as an absolute opposing category to the expert states during training, we further propose a novel calibrated supervised contrastive loss function, denoted as $\mathcal{L_{\text{C-SupCon}}}$, which considers the agent states as a mixture of both expert and agent states. The training framework of CAIL is shown in Figure \ref{fig:CAIL}. 

\subsection{Contrastive Representation on Visual States}
\noindent\textbf{Unsupervised Contrastive Loss.}
Prior research in GAIL has focused only on the comparison between expert and agent states while ignoring the significant number of available agent states in the replay buffer. One potential approach to leverage the agent states via contrastive learning is to employ unsupervised contrastive learning with the large unlabeled agent state dataset. In practice, we obtain $2N$ augmented agent states $\bm v^a:=\{\bm v^a_i\}_{i=1}^{2N}$ from a batch of $N$ agent states. Two augmentations from the same source are denoted as $(\bm v^a_i,p(\bm v^a_i))$, where $p(\bm v^a_i)$ is one of the two augmented states other than $\bm v^a_i$. Since the labeling information is unavailable, we consider $p(\bm v^a_i)$ as the sole positive example to $\bm v^a_i$, and we aim to minimize the distance between the representations of $\bm v^a_i$ and $p(\bm v^a_i)$. The remaining $2N-2$ augmented states in the same batch are regarded as negatives to $\bm v^a_i$ and their representations are expected to be far from $\bm v^a_i$. We adopt the InfoNCE loss~\cite{oord2018representation}, which is widely-accepted in unsupervised contrastive learning to formulate our unsupervised contrastive loss for agent states $\bm v^a_i$:
\begin{small}
\begin{align}
    &\ell^{\text{InfoNCE}}(\bm v^a_i, \bm v^{a+}_i, \bm v^{a-}, f, h) \notag\\ 
    &= -\log \frac{\exp\left(sim\left(h\left(f\left(\bm v^a_i\right)\right), h\left(f\left(\bm v^{a+}_i\right)\right)\right)/\tau\right)}{\sum_{j} \exp\left(sim\left(h\left(f\left(\bm v^a_i\right)\right), h\left(f\left(\bm v^{a-}_j\right)\right)\right)/\tau\right)}~,\label{eq:InfoNCE} 
\end{align}
\end{small}
where the $sim(\cdot)$ denotes the cosine similarity score function, $\bm v^{a+}_i$ is the positive states pair with $\bm v^a_i$, $\bm v^{a-}$ is the contrastive states set, $h$ is the MLP projection head and $\tau$ is the temperature parameter. The minimization of $\ell^{\text{InfoNCE}}$ results in the pulling together of representations of $v^{a+}_i$ and $\bm v^a_i$, while those of dissimilar states ($\bm v^{a-}$) are pushed away. Our unsupervised contrastive loss $\mathcal{L_{\text{UnSupCon}}}$ is defined as the average $\ell^{\text{InfoNCE}}$ loss over all $2N$ state augmentations in the batch and denoted as:
\begin{small}
\begin{equation}
\label{eq:UnSupCon loss}
\mathcal{L_{\text{UnSupCon}}} = \frac{1}{2N} \sum_{i=1}^{2N} \ell^{\text{InfoNCE}}(\bm v_i^a, p(\bm v_i^a),\bm v^a \setminus \bm v^a_i, f, h_{\text{unsup}}).~
\end{equation}
\end{small}
Here, $h_{\text{unsup}}$ is a dedicated projection head used in unsupervised contrastive loss. Note that the training process is identical to SimCLR~\cite{chen2020simple} without the following Supervised Contrastive Loss.

\noindent\textbf{Supervised Contrastive Loss for Discrimination.}
Unsupervised contrastive loss exploits sufficient agent states with replay buffer to learn a discriminative representation, however we are also expecting the learned representation is distinguishable between agents and experts. In practice, expert states could have regular patterns and strong clustering effect, while mistakes in agent states often differ from one another. Therefore, a good representation should push expert states together while pulling away expert states with those agent states. This can be achieved by utilizing supervised contrastive loss~\cite{khosla2020supervised}. While the supervised contrastive loss was designed for supervised settings, we can treat the expert states and agent states as distinct groups, allowing us to utilize it as a binary classification problem for training the encoder in an unsupervised setting.

Unlike unsupervised contrastive loss, which employs only one positive example, supervised contrastive loss generalizes $\mathcal{L_{\text{UnSupCon}}}$ to any number of positive examples. Specifically, in our approach, we consider all augmented expert states $\bm v^e=\{\bm v_i^e\}_{i = 1}^N$ in one batch as states of the same class, while all augmented agent states $\bm v^a$ are viewed as representations of the opposite class. We denote all augmented states in one batch as $\bm v = \bm v^a \cup \bm v^e$. The supervised contrastive loss for each state $\bm v^e_i \in \bm v^e$ is defined as follows,
\begin{small}
\begin{align}
    &\ell^{\text{Sup}}(\bm v^e_i, \bm v^e) \notag\\&= -\frac{1}{|\bm v^e|-1}\sum_{\bm v^+_j \in \bm v^e\setminus \bm v^e_i}\ell^{\text{InfoNCE}}(\bm v_i^e, \bm v_j^+, \bm v\setminus\bm v_i^e, f, h_{\text{sup}})~
    \label{eq:sup-con single for expert}
\end{align}
\end{small}
where $h_{\text{sup}}$ is the projection head dedicated for supervised contrastive loss and $|\bm v^e|$ is the cardinality of $\bm v^e$. 
While the expert 
The final supervised contrastive loss for the CAIL is the average supervised contrastive loss over all expert states, denoted as,
\begin{equation}
\label{eq:SupCon loss}
\mathcal{L_{\text{SupCon}}} = \frac{1}{N} \sum_{i=1}^N \ell^{\text{Sup}}(\bm v^e_i, \bm v^e).
\end{equation}
Through the utilization of the supervised objective $\mathcal{L_{\text{SupCon}}}$, the encoder is capable of extracting a representation that is not only consistent with $\mathcal{L_{\text{UnSupCon}}}$, but also has an enhanced ability to discriminate between experts and agents.

\subsection{Calibrating the Contrastive Representation}
In contrast to vanilla supervised contrastive learning, where label information is pre-determined and fixed, our approach involves training an agent that progressively acquires expertise over time, particularly in the latter stages of training~\cite{wang2023unlabeled}. At this point, some of the well-trained agent states may even be more realistic than the expert. Thus, continuously treating them as opposite examples to experts could lead to unstable training and may be deemed unfair. Instead, it is more reasonable to regard the agent states as a mixture of positive states and negative states towards expert states. In this way, we are able to better account for the varying levels of expertise exhibited by the agent states and enhance the stability and fairness of the training process. 

However, determining which states are well-trained and which require further training can be a challenging task. To address this issue, we consider the agent state $\bm v^{a}_i$ as a high-quality sample (\textit{e.g.}, expert state) with a probability of $\alpha$, and as a low-quality sample with a probability of $1-\alpha$. When treating $\bm v^{a}_i$ as high-quality sample, we utilize the supervised contrastive loss $\ell^{\text{Sup}}$ to train the encoder. In this case, all expert states $\bm v^e_i$ and the corresponding augmented state $p(\bm v^{a}_i)$ are treated as positive pairs for $\bm v^{a}_i$. Conversely, when treating agent samples as low-quality samples, we change the training objective to the unsupervised loss $\ell^{\text{InfoNCE}}$. Here, only the augmented state $p(\bm v^{a}_i)$ is treated as a positive pair for $\bm v^{a}_i$, while all other states are considered as negatives. We then obtain a calibrated supervised contrastive loss, which can be expressed as:
\begin{align}
\label{eq:C-SupCon loss}
&\mathcal{L_{\text{C-SupCon}}} =  \alpha \mathbb{E}_{\bm v^a_i\sim\rho_{\pi_\theta}}\left[\ell^{\text{Sup}}(\bm v^a_i, \bm v^e \cup \{v^a_i, p(v^a_i)\})\right] \\
&+  (1-\alpha)\mathbb{E}_{\bm v^a_i\sim\rho_{\pi_\theta}}\left[\ell^{\text{InfoNCE}}(\bm v_i^a, p(\bm v_i^a),\bm v \setminus \bm v^a_i, f, h_{\text{sup}})\right]\notag
\end{align}
Combining Eq. (\ref{gail_df}), Eq. (\ref{eq:UnSupCon loss}) and Eq. (\ref{eq:C-SupCon loss}), we can obtain the final training objective of CAIL as:
\begin{equation}
\label{eq:training objective CAIL}
\mathcal{L_{\text{CAIL}}} = \mathcal{L_{\text{dis}}} + \lambda_1\mathcal{L_{\text{UnSupCon}}} + \lambda_2\mathcal{L_{\text{C-SupCon}}}
\end{equation}
where $\lambda_1$ and $\lambda_2$ are the hyperparameters to control the contrastive loss ratio and $\mathcal{L_{\text{dis}}} = \frac{1}{N}\sum_{i=1}^{N}-\log(h_d(r^e_i))-\log(1-h_d(r^a_i))$ in practice. In this study, we keep both $\lambda_1$ and $\lambda_2$  equal to $1$.

\subsection{Overall Algorithm}
The whole algorithm is summarized in Algorithm \ref{al1}. The update of the discriminator $D\rightarrow f\circ h_d$ has been introduced in the above subsection. The policy learning part in CAIL is based on the off-policy reinforcement learning method DDPG~\cite{silver2014deterministic}. In DDPG, the policy network $\pi_\theta$ and the critic $Q_{\phi_{1}}$ is optimized alternatively. To relieve the overestimation during the critic learning, another critic network $Q_{\phi_{2}}$ is introduced following the idea of double Q-learning~\cite{van2016deep}.

\noindent\textbf{Value Iteration.}
The value iteration process in DDPG basically follows the Deep Q-Learning (DQN), which is a boosting method that updates the critic networks $Q_{\phi}$ based on its own predicted Q-value at the next state. The parameter $\phi$ is learned by minimizing the Temporal-Difference (TD) error,
\begin{align}
    \mathcal{L}(\phi, \mathcal{B})=\mathbb{E}_{t\sim \mathcal{B}}\big[(Q_{\phi_i}(v, a) - (r+\gamma(1-d)\Gamma))^2\big],     \label{eq:bellman}\\\notag \forall i\in \{1,2\}
\end{align}
where $\mathcal{B}$ denotes the replay buffer, $t=(v,a,r, v^\prime,d)$ is a tuple with visual state $\textbf{v}$, action $a$, next visual state $v^\prime$ and done signal $d$. The reward $r$ is provided by discriminator in adversarial imitation learning. The estimation of next state's Q-value $\Gamma$ is defined as $\Gamma=(\min_{i=1,2} Q_{\overline{\phi}_i}(v^\prime, a^\prime))$, where $a^\prime \sim \pi_\theta(a|s^\prime)$ and $\overline{\phi}$ is an exponential moving average of the weights.

\noindent\textbf{Policy Iteration.} 
Since the critic $Q_\phi$ has been learned to indicate the quality of the action, the policy network $\pi_\theta$ thus should be trained to maximize the expected return as
\begin{align}
    \mathcal{L}(\theta) = - \mathbb{E}_{(v,a)\sim \pi_\theta}\Big[\min_{i=1,2} Q_{\phi_i}(v,a)\Big],
    \label{eq:actor}
\end{align}
where action $a=\pi_\theta(v)+\epsilon$, $\epsilon\sim \text{clip}(\mathcal{N}(0,\sigma^2), -c, c)$ and $\sigma$ denotes the exploration noise. Following common practice, we do not use actor gradient to update the parameter of the image encoder $f$.
\begin{algorithm}[!t] 
	\caption{\textbf{C}ontrastive \textbf{A}dversarial \textbf{I}mitation \textbf{L}earning} 
	\label{alg:Framwork} 
	\begin{algorithmic}[1]
		\REQUIRE ~~\\
            Empty replay buffer $\mathcal{B}$, Image encoder $f$, MLP trunk $d_h$, Policy network $\pi_\theta$, Expert observations $\mathcal{D}_e$;
		\STATE Initialize $f$, $d_h$, and $\pi_{\theta}$;
		\FOR{iter = 0, 1, 2, ...}
            \STATE Generate an episode $\{\textbf{v}_1, \textbf{a}_1, \textbf{v}_2, \textbf{a}_2, ...\}$ with $\pi_\theta$
            \STATE Store $\{(\textbf{v}_t, \textbf{a}_t, \textbf{v}_{t+1})\}$ to replay buffer $\mathcal{B}$
            \STATE Sample $\textbf{v}^a\sim\mathcal{B}$, $\textbf{v}^e\sim\mathcal{D}_e$
            \STATE Update the discriminator formed by image encoder $f$ and MLP trunk $h_d$ with Eq. (\ref{eq:training objective CAIL})
		\STATE Update $\pi_\theta$ by off-policy RL method with Eq. (10)          
		\ENDFOR
		\label{algo1}
	\end{algorithmic}
	\label{al1}
\end{algorithm}

\subsection{Convergence Analysis}
We provide convergence analysis on the objective function of CAIL since it adds contrastive regularizer term into the framework of GAIL. We first show that the objective function of CAIL has the same convergence point as GAIL in Theorem \ref{theo1}. Then, we provide analysis on the update of encoder $f$ and how the contrastive loss helps to improve the performance.
\begin{theorem}
    Rewrite the objective function of CAIL as $\min_{\theta,f}\max_{h_d} \mathcal{J}_{h_d,f}(\theta)$, where
    \begin{align}
        \mathcal{J}_{h_d,f}(\theta)=&\mathbb{E}_{\bm v\sim\rho_{\pi_e}} [\log h_d(f(\bm v))] \\ \nonumber &-\mathbb{E}_{\bm v\sim\rho_{\pi_\theta}} [\log(1-h_d(f(\bm v)))] + \Psi(f),
    \end{align}
    where $\Psi(f)$ denotes the contrastive constraint. Regardless of the encoder $f$, $\mathcal{L}_{h_d,f}(\theta)$ can converge with respect to $\theta$, and the convergence point is reached when $\rho_{\pi_{\theta^\ast}}=\rho_{\pi_e}$.
    \label{theo1}
\end{theorem}
The proof of Theorem \ref{theo1} is available in the supplementary material. As stated in the theorem, $\mathcal{L}_{h_d,f}(\theta)$ can converge to $\mathcal{L}_{h_d,f}(\theta^\ast)$ when given a fixed encoder $f$. Since the contrastive constraint $\Psi(f)$ only applies on $f$, they will not affect the convergence of agent policy $\pi_\theta$. While adversarial IL often faces the challenge of unstable training and hard convergence, the contrastive constraint helps to improve the ability of representation in discriminator, which makes the training more stable and converge fast. Empirical results also identify the sample-efficient property of CAIL.

\begin{figure}[!h]
    \centering
    \includegraphics[width=3.2in]{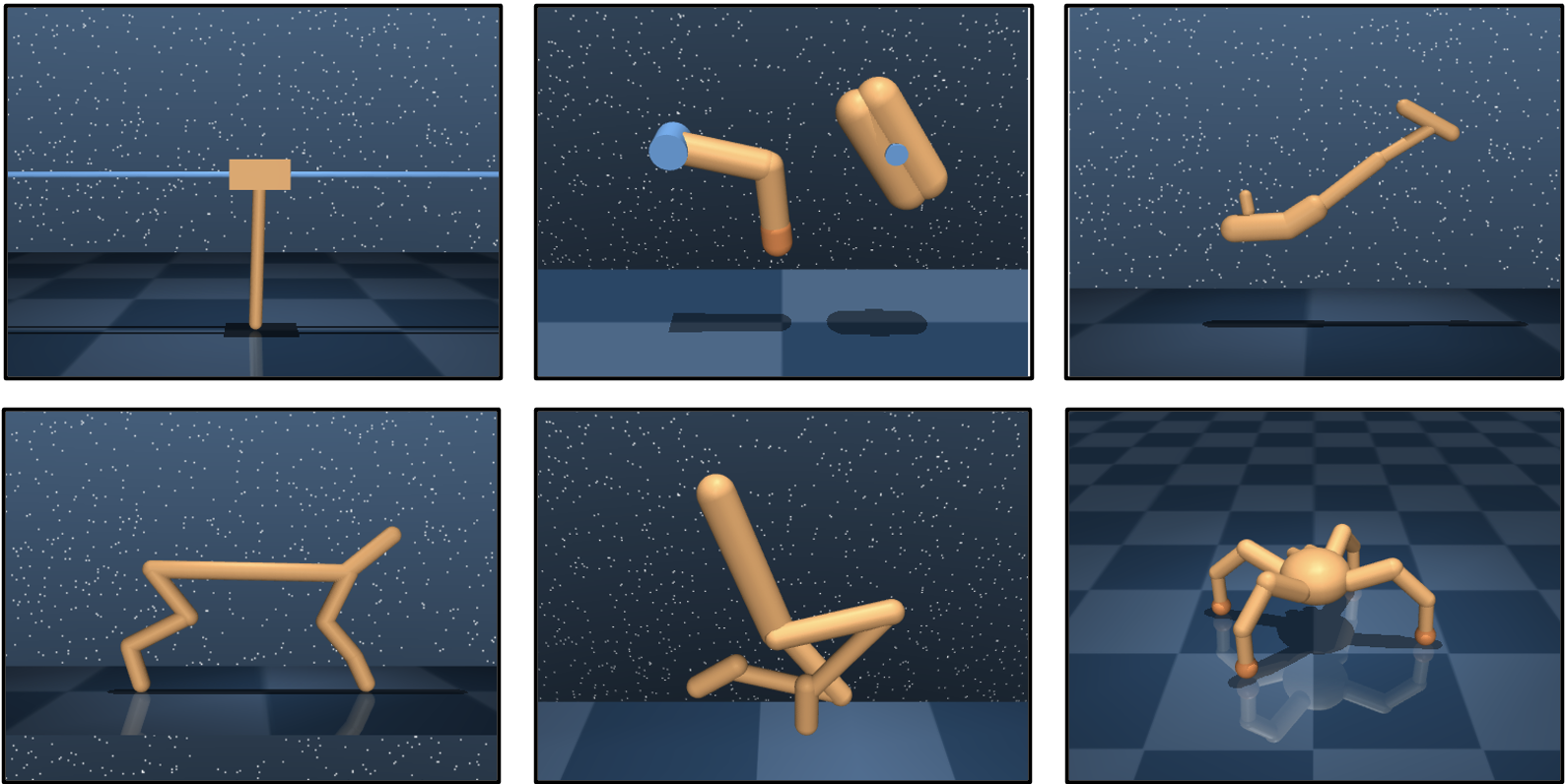}
    \caption{Benchmarking domains. \textbf{Top}: Cartpole, Finger, Hopper. \textbf{Bottom}: Cheetah, Walker, and Quadruped.}
    \label{a}
    \vskip -0.1in
\end{figure}
\section{Experiments}
In this section, we conduct experiments to verify the effectiveness of CAIL from different aspects. Evaluations are conducted on pixel-based benchmark tasks on DeepMind Control Suit (DMControl)~\cite{tunyasuvunakool2020dm_control}. 

\noindent\textbf{Setups.}
We evaluate CAIL on 9 MuJoCo~\cite{todorov2012mujoco} tasks in the DMControl Suite. These tasks include Cartpole Swingup, Finger Spin, Cheetah Run, Hopper Hop, Hopper Stand, Walker Stand, Walker Walk, Walker Run and Quadruped Run. As for expert demonstrations, we use the public dataset in ROT~\cite{haldar2022watch}. More details can be found in the supplementary material.
\begin{table*}[!tbp]
\small
    \centering
    \caption{Performance of CAIL and compared methods in 9 DeepMind Control tasks at both sample-efficient \textit{500K} and \textit{1M} steps. The agent is measured by the average and standard variance of rewards along 10 trajectories (\textit{i.e.}, the higher the better). We repeat the experiments for 5 trials with different random seeds. In contrast to CAIL, CAIL(w/o cal) replaces the calibrated supervised contrastive loss with plain one.}
    \setlength{\tabcolsep}{3.75mm}{
    \begin{tabular}{l|ccccc|cc}
    \hline
    \textit{500K Steps} & GAIL & GAIL-SE & PCIL & CAIL(w/o cal) & CAIL & BC & Expert  \\ 
    \hline
      Cartpole Swingup  & 186$\pm$20 & 734$\pm$160  & 296$\pm$52 & 790$\pm$61  & \bf 838$\pm$4 & 521$\pm$120 & 859$\pm$0 \\
      Finger Spin       & 0$\pm$0    & 303$\pm$191  & 408$\pm$297& 460$\pm$193 & \bf 642$\pm$186 & 284$\pm$120 & 976$\pm$9 \\
      Cheetah Run       & 69$\pm$27  & 514$\pm$33   & 461$\pm$51 & 497$\pm$44  & \bf 538$\pm$34 & 185$\pm$49  & 890$\pm$19 \\
      Hopper Hop        & 10$\pm$7   & 8$\pm$8      & 38$\pm$18  & 51$\pm$23   & \bf 73$\pm$12  & 109$\pm$18 & 318$\pm$7 \\
      Hopper Stand      & 5$\pm$3    & 270$\pm$343  & 451$\pm$199& 290$\pm$306 & \bf 541$\pm$305& 386$\pm$72  & 976$\pm$9 \\
      Walker Stand      & 272$\pm$95 & 513$\pm$286  & 930$\pm$18 & 910$\pm$44  & \bf 961$\pm$9  & 496$\pm$70  & 939$\pm$9 \\
      Walker Walk       & 69$\pm$49  & 268$\pm$61   & 203$\pm$19 & 350$\pm$136 & \bf 463$\pm$126& 556$\pm$110 & 970$\pm$20 \\
      Walker Run        & 24$\pm$6   & 57$\pm$17    & 146$\pm$10 & 148$\pm$28  & \bf 157$\pm$9  & 378$\pm$82 & 778$\pm$10 \\
      Quadruped Run     & 151$\pm$57 & 212$\pm$5   & 228$\pm$119& 238$\pm$50  & \bf 306$\pm$110& 277$\pm$58  & 547$\pm$136 \\
        \hline
    \textit{1M Steps} & GAIL & GAIL-SE & PCIL & CAIL(w/o cal) & CAIL & BC & Expert  \\ 
        \hline
      Cartpole Swingup  & 199$\pm$17 & 801$\pm$91 & 67$\pm$36   & 687$\pm$312 & \bf 837$\pm$23   & 521$\pm$120 & 859$\pm$0 \\
    Finger Spin       & 0$\pm$0    & 407$\pm$250 & 534$\pm$233 & 704$\pm$122 & \bf 785$\pm$99 & 284$\pm$120 & 976$\pm$9 \\
      Cheetah Run       & 84$\pm$30  & 624$\pm$35 & 662$\pm$27  & 689$\pm$37  & \bf 725$\pm$31 & 185$\pm$49  & 890$\pm$19 \\
      Hopper Hop        & 0$\pm$0    & 121$\pm$24  & 158$\pm$8   & \bf  184$\pm$25  & 182$\pm$20 & 109$\pm$18 & 318$\pm$7 \\
      Hopper Stand      & 5$\pm$3    & 747$\pm$63 & 733$\pm$104 & 754$\pm$106 & \bf 777$\pm$42 & 386$\pm$72  & 976$\pm$9 \\
      Walker Stand      & 275$\pm$100& 764$\pm$251 & 827$\pm$38  & \bf 859$\pm$101 & 831$\pm$154 & 496$\pm$70  & 939$\pm$9 \\
      Walker Walk       & 63$\pm$34  & \bf 953$\pm$3 & 952$\pm$0   & 940$\pm$13  &  938$\pm$6  & 556$\pm$110 & 970$\pm$20 \\
      Walker Run        & 28$\pm$8   & 133$\pm$28  & 519$\pm$59  & 468$\pm$66  & \bf 526$\pm$48 & 378$\pm$82 & 778$\pm$10 \\
      Quadruped Run     & 115$\pm$60 & 296$\pm$82  & \bf 427$\pm$35  & 322$\pm$38  & 382$\pm$27 & 277$\pm$58 & 547$\pm$136 \\
        \hline
    \end{tabular}
    }
    \label{tab:main}
\end{table*}
\begin{figure*}[!t]
    \includegraphics[width=7in]{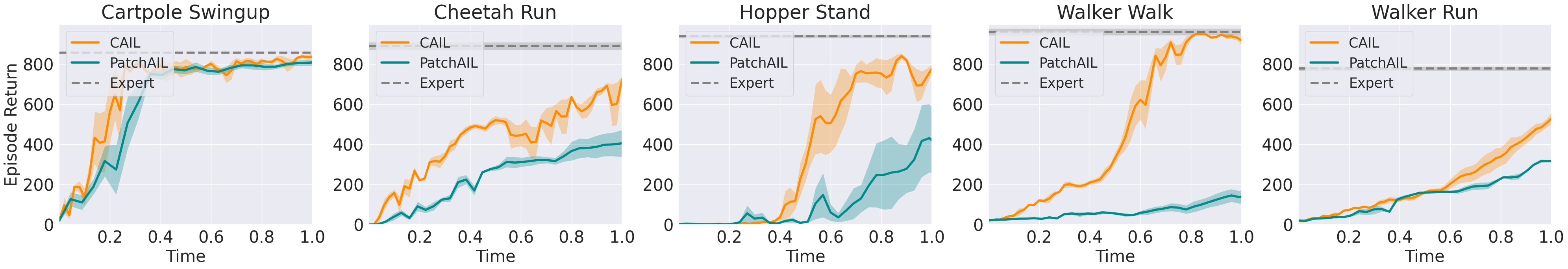}
    \caption{Learning curves of CAIL and PatchAIL on 5 DMC tasks with respect to training time. We scale the training time and `1' denotes the time that CAIL completed \textit{1M} steps. It is obvious that given the same training time, CAIL can outperform PatchAIL.}
    \label{fig:patchirl}
\end{figure*}

\noindent\textbf{Compared Methods.}
We compare CAIL with the other five methods: Behavioral Cloning (BC), GAIL, GAIL with Shared-Encoder (GAIL-SE)~\cite{cohen2021imitation}, PCIL~\cite{huang2023policy} and PatchAIL~\cite{liu2023visual}.
BC learns a policy that maps visual states to actions via supervised learning.  GAIL adopts the policy network and the discriminator as two independent models, while GAIL-SE makes both policy network and discriminator network share a common image encoder. PCIL also utilizes the idea of contrastive learning to enhance the representation. PatchAIL exploits a patch discriminator during training and produces patch rewards for the agent.

\subsection{Imitation Performance}
The results are presented in Table \ref{tab:main}. We can observe that GAIL struggles to achieve desirable results in visual imitation learning tasks, even in the simplest task Cartpole Swingup. For tasks like Hopper Stand, the agent is not even able to learn a better policy than random. These results align with previous research studies conducted in~\cite{tucker2018inverse}. By adopting a shared image encoder for both policy and discriminator, the agent trained by GAIL-SE achieves at least double the reward as plain GAIL in all nine tasks. However, there is still a notable gap in comparison to expert performance. PCIL and CAIL improve the performance over GAIL baseline greatly, which suggests that incorporating contrastive representation enables the agent to converge to a better solution. Our proposed method CAIL performs best at \textit{500K} steps in all 9 tasks, which demonstrate the sample-efficient property of CAIL. At \textit{1M} steps, CAIL and CAIL (w/o cal) demonstrate best performance among 7 out of 9 tasks.

\begin{figure}[!t]
    \centering
    \includegraphics[width=3.2in]{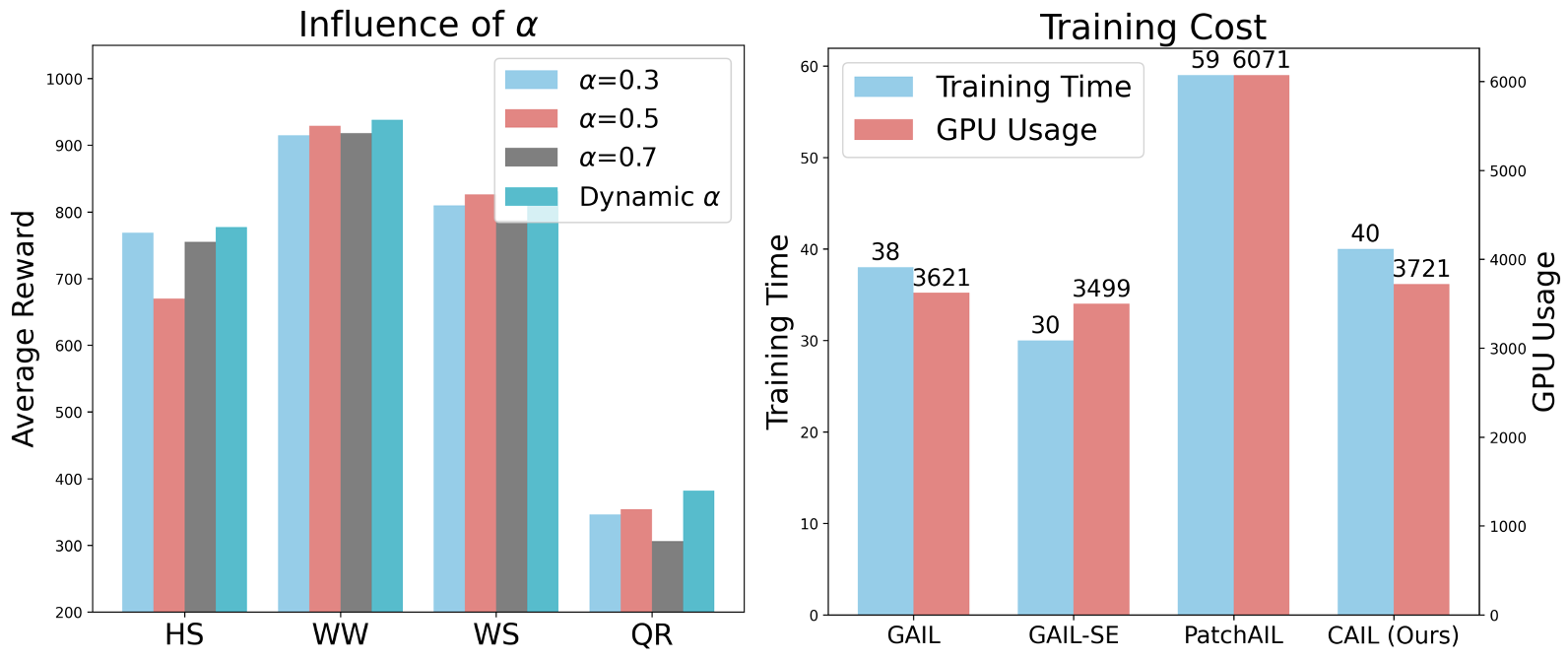}
    \caption{\textbf{Left:} Impact of $\alpha$. `HS', `WS', `WW' and `QR' denote `Hopper Stand', `Walker Stand', `Walker Walk' and `Quadruped Run' respectively. \textbf{Right:} Training cost of CAIL and compared methods measured by GPU memory usage and training time.}
    \label{fig:label}
\end{figure}

\noindent\textbf{Compare with PatchAIL.} PatchAIL is a state-of-the-art visual IL method that adopts a patch discriminator in the adversarial training. While it exhibits impressive asymptotic performance, it comes with a substantial increase in training cost compared to other methods. In Figure \ref{fig:label}, we present the training time and GPU memory usage of PatchAIL in the Cartpole Swingup task. It shows that PatchAIL incurs a significant overhead, with both training time and GPU memory usage nearly doubling in comparison to GAIL. We provide the learning curves of CAIL and PatchAIL under the same training time budget in Figure \ref{fig:patchirl}, which shows that CAIL is significantly computational efficient than PatchAIL. We defer the asymptotic performance of CAIL and PatchAIL to the supplementary material.

\begin{figure}[!t]
    \centering
    \includegraphics[width=3.2in]{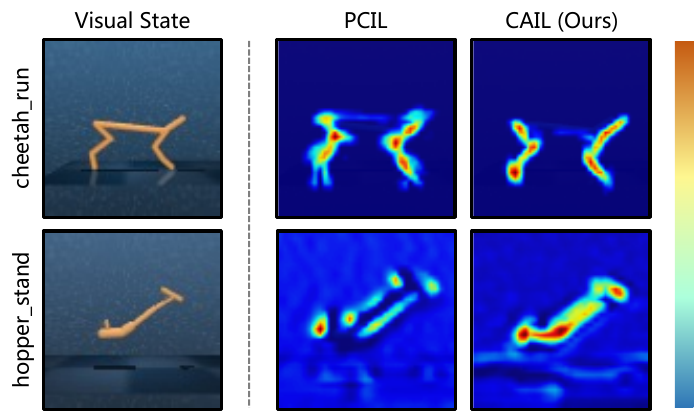}
    \caption{
    Spatial attention map of discriminator at $\textit{1M}$ steps. The map shows the region that the discriminators focus on to make the decision. }
    \vskip -0.05in
    \label{fig:cam_map}
\end{figure}

\noindent\textbf{Visualization.} As shown in Figure \ref{fig:cam_map}, we visualized the spatial attention map of the discriminator in two tasks by Grad-CAM~\cite{selvaraju2017grad}. We compare CAIL with its closest baseline PCIL. The attention map visually highlights the areas that the discriminator prioritizes in decision-making, with the red region indicating a more significant influence on the decision. From the figure, it is evident that the attention map of both methods can focus on the joint of agent's body, which is intuitively important to distinguish different visual states. Compared to PCIL, the attention map of CAIL has a better coverage on the joint of agent's body in its attention map. This visual interpretation suggests that CAIL has a better ability in capturing the differences among different visual states, thus makes it achieve better performance.



\noindent\textbf{Impact of $\alpha$.}
The parameter $\alpha$ represents prior information indicating the probability of the agent's demonstration being `positive' during training. In this study, we examine how different values of $\alpha$ can affect the performance of the calibrated contrastive loss in the experiment. Intuitively, at the beginning of training, the agent demonstrations are more likely to be `negative' samples. As training progresses, the probability of the agent demonstration being `positive' increases accordingly. Hence, it is unsuitable to regard the agent demonstrations as `negative' samples. To address this problem, we design a dynamic $\alpha$ that increases from 0.3 to 0.5 linearly during training. The results of dynamic $\alpha$ are presented in Figure \ref{fig:label}. We observe that CAIL is relatively tolerant of different $\alpha$ values. Generally, using a dynamic $\alpha$ leads to better performance in most cases, which aligns with our hypothesis. We then apply the dynamic $\alpha$ to CAIL, which can also be considered a method that does not require prior information.

\subsection{Empirical Study on Contrastive Loss}

\noindent\textbf{Data Augmentation v.s. Contrastive Learning.}
Some existing works find that directly using strong augmentation in model training can cause instability and hurt the performance of the model~\cite{jeong2021training}. 
To verify this idea, we apply the same data augmentation in GAIL-SE. The results are shown in Table \ref{tab:contravsaug}. Except for the Walker Stand task, we observe a fact that adding augmentations into GAIL-SE is not very helpful to improve performance since it leads to an average of 40.4\% drops compared to the baseline. By contrast, we find the CAIL outperforms the baseline in these 5 tasks with an 47.2\% improvement. Hence, we conclude that using our calibrated contrastive loss incorporated with data augmentation is better than directly applying data augmentation in adversarial imitation learning.
\begin{table}[t]
\small
    \centering
    \caption{The comparison between data augmentation and contrastive learning in GAIL-SE. 
    }
    \setlength{\tabcolsep}{0.7mm}{
    \begin{tabular}{lccc}
        \hline
        \multirow{2}{*}{\textit{500K Steps}}  & GAIL-SE & GAIL-SE & CAIL \\ 
                                             & w/o Aug & w/ Aug & (Ours) \\
        \hline
        Cartpole Swing      & 734  & 729$\pm$40($\downtri$0.1\%) & \textbf{838$\pm$4} ($\uptri$14.2\%) \\
        Cheetah Run         & 514  & 533$\pm$34($\uptri$3.7\%) & \textbf{538$\pm$34}($\uptri$4.7\%)\\
        Hopper Stand        & 270  & 206$\pm$89($\downtri$23.3\%) & \textbf{541$\pm$305}($\uptri$100\%)\\
        Walker Stand        & 513 & 881$\pm$15($\uptri$71.7\%) & \textbf{961$\pm$9}($\uptri$87.3\%)\\
        Walker Walk         & 268  & 200$\pm$5($\downtri$25.4\%) & \textbf{463$\pm$126}($\uptri$72.8\%)\\
        Quadruped Run       & 212   & 222$\pm$74($\uptri$4.7\%) & \textbf{306$\pm$110}($\uptri$44.3\%)\\
    \hline
    \end{tabular}
    }
    \label{tab:contravsaug}
\end{table}

\noindent\textbf{Different Augmentation Ways.}
In our experiments, we utilize ``Random-Shift'' augmentation for producing different views of demonstrations, which has been shown to be the most useful data augmentation technique in pixel-based reinforcement learning problems~\cite{laskin2020reinforcement}. To verify how other augmentation ways (\textit{e.g.}, Random Crop, Random Cutout and Random Aug) may affect the performance of adversarial imitation learning, we provide results in the Hopper Stand task in Table \ref{tab:aug}. We find that CAIL can outperform GAIL-SE with all 4 different augmentations ways, and using ``Random-Shift'' augmentation results in the best performance. We also observe the fact that using some augmentations ways can degrade the performance of AIL (\textit{e.g.}, Random Cutout and Random Aug).
\begin{table}[t]
\small
    \centering
    \caption{The performance of GAIL-SE and CAIL when using different data augmentation ways in the `HS' task.}
    \setlength{\tabcolsep}{6.5mm}{
    \begin{tabular}{lcc}
    \hline
        \textit{Augmentation} & GAIL-SE  & CAIL \\ 
            \hline
        No Aug       & 602$\pm$163  &  N/A   \\
        +Random Shift       & 676$\pm$130  & \bf 777$\pm$42  \\
        +Random Crop        & 649$\pm$208  & \bf 755$\pm$71  \\
        +Random Cutout      & 530$\pm$260  & \bf 691$\pm$92  \\
        +Random Aug         & 524$\pm$165  & \bf 549$\pm$115      \\
        \hline
    \end{tabular}
    }
    \label{tab:aug}
\end{table}

\section{Conclusion}
In this work, we propose a novel solution to address the challenge of visual adversarial imitation learning using contrastive learning. In high-dimensional visual states, it is difficult to capture changes in the agent's behavior as easily as in low-dimensional states. Therefore, a good representation is essential to ensure the performance of visual imitation learning. Existing solutions either attempt to separate the process of learning representation and decision-making or adopt complex architectures to enhance the representation. In contrast, we introduce contrastive learning to improve AIL's ability to discriminate between different demonstrations. Specifically, three different contrastive losses are adopted to learn a good image encoder for AIL. We evaluate the effectiveness of the proposed method on the DMControl suite, and empirical results demonstrate its superiority from various aspects.

\bibliographystyle{named}
\bibliography{ijcai24}

\end{document}